\documentclass[runningheads]{llncs}
\usepackage[T1]{fontenc}
\usepackage[utf8]{inputenc} 
\usepackage{graphicx}
\usepackage{graphics} 
\usepackage{epsfig} 
\usepackage{mathptmx} 
\usepackage{times} 
\usepackage{amsmath} 
\usepackage{amssymb}  
\usepackage{bm}
\usepackage{eucal}
\usepackage{subcaption}
\usepackage{setspace}
\usepackage{diagbox} 
\usepackage{todonotes}
\usepackage{multicol, blindtext}
\usepackage{multirow}
\usepackage{algorithm}
\usepackage{algpseudocode}
\usepackage[font=small,labelfont=bf]{caption}
\usepackage{changes} 
\newcommand{\HEADER}[1]{\ALC@it\underline{\textsc{#1}}\begin{ALC@g}}
	\newcommand{\ENDHEADER}{\end{ALC@g}}
\makeatother
\usepackage[pagebackref=true,breaklinks=true,colorlinks=true,bookmarks=false]{hyperref}
\usepackage{xcolor}
\definecolor{ForestGreen}{RGB}{1,184,84}
\definecolor{Burgund}{RGB}{200,1,1}
\newcommand{\ie}{\textit{i}.\textit{e}.}
\newcommand{\eg}{\textit{e}.\textit{g}.}
\newcommand{\etal}{\textit{et~al}.}
\newcommand{\wrt}{\textit{w}.\textit{r}.\textit{t}.}

\newcommand{\cflow}{\emph{c-flow}\xspace}

\newcommand\copyrighttext{%
	\footnotesize © Submitted to SafeCOMP 2024. Personal use of this material is permitted. Permission must be obtained for all other uses, in any current or future media, including
	reprinting/republishing this material for advertising or promotional purposes, creating new collective works, for resale or redistribution to servers or lists,
	or reuse of any copyrighted component of this work in other works.
}
\newcommand\copyrightnotice{%
	\begin{tikzpicture}[remember picture,overlay]
	\node[anchor=north,yshift=-10pt] at (current page.north) {\fbox{\parbox{\dimexpr\textwidth-\fboxsep-\fboxrule\relax}{\copyrighttext}}};
	\end{tikzpicture}%
}

\usepackage[shrink=65, selected=True]{microtype}
\begin{document}
%
\title{A Flow-based Credibility Metric for Safety-critical Pedestrian Detection}
%
%
\author{Maria Lyssenko\inst{1,2}\and
Christoph Gladisch\inst{1}\and
Christian Heinzemann\inst{1}\and
Matthias Woehrle \inst{1}\and
Rudolph Triebel \inst{3,4}}
\authorrunning{M. Lyssenko~\etal}
\institute{Robert Bosch GmbH, Corporate Research, Germany
	\email{firstname.lastname@de.bosch.com}\\
	\and Technical University of Munich, Germany
	\email{maria.lyssenko@tum.de}\\
	\and German Aerospace Center, Germany
	\email{rudolph.triebel@dlr.de}\\
	\and Karlsruhe Institute of Technology, Germany\\
	\email{rudolph.triebel@kit.edu}\\}

\maketitle              
\copyrightnotice

\begin{abstract}
Safety is of utmost importance for perception in automated driving (AD). However, a prime safety concern in state-of-the art object detection is that standard evaluation schemes utilize safety-agnostic metrics to argue sufficient detection performance. Hence, it is imperative to leverage supplementary domain knowledge to accentuate safety-critical misdetections during evaluation tasks. To tackle the underspecification, this paper introduces a novel credibility metric, called \cflow, for pedestrian bounding boxes. To this end, \cflow relies on a complementary optical flow signal from image sequences and enhances the analyses of safety-critical misdetections without requiring additional labels. We implement and evaluate \cflow with a state-of-the-art pedestrian detector on a large AD dataset. Our analysis demonstrates that \cflow allows developers to identify safety-critical misdetections.
\end{abstract}

\section{Introduction}

In automated driving (AD), safety is a crucial aspect of perception systems.
Driven by the remarkable performance demonstrated in perception tasks such as camera-based object detection, the demand for systems utilizing deep neural networks (DNN) has surged in the field of AD.
In respect thereof, ensuring accurate and reliable perception of vulnerable road users (VRU) like pedestrians is a significant requirement.

Hence, for a systematic safety argumentation the consideration of safety concerns is of utmost importance~\cite{Abrecht2023DeepLS}.
Among other things, the concerns encompass the ubiquitous safety-agnostic evaluation that treats all misdetections equally irrespective of their particular relevance for the safe driving task.
To tackle safety-agnosticism, recent works by Wolf~\etal~\cite{WolfDE21}, Ceccarelli~\etal~\cite{Ceccarelli2022EvaluatingO}, and ourselves~\cite{LGH+21,LGH+22} leverage domain knowledge (in form of, \eg~temporal distance) to incorporate a notion of safety into the evaluation of object detectors.
This safety-awareness prioritizes pedestrians that are especially relevant to the autonomous vehicle (AV) for further downstream tasks like planning and control.

\begin{figure}[t!]
	\centering
	\includegraphics[width=0.7\columnwidth]{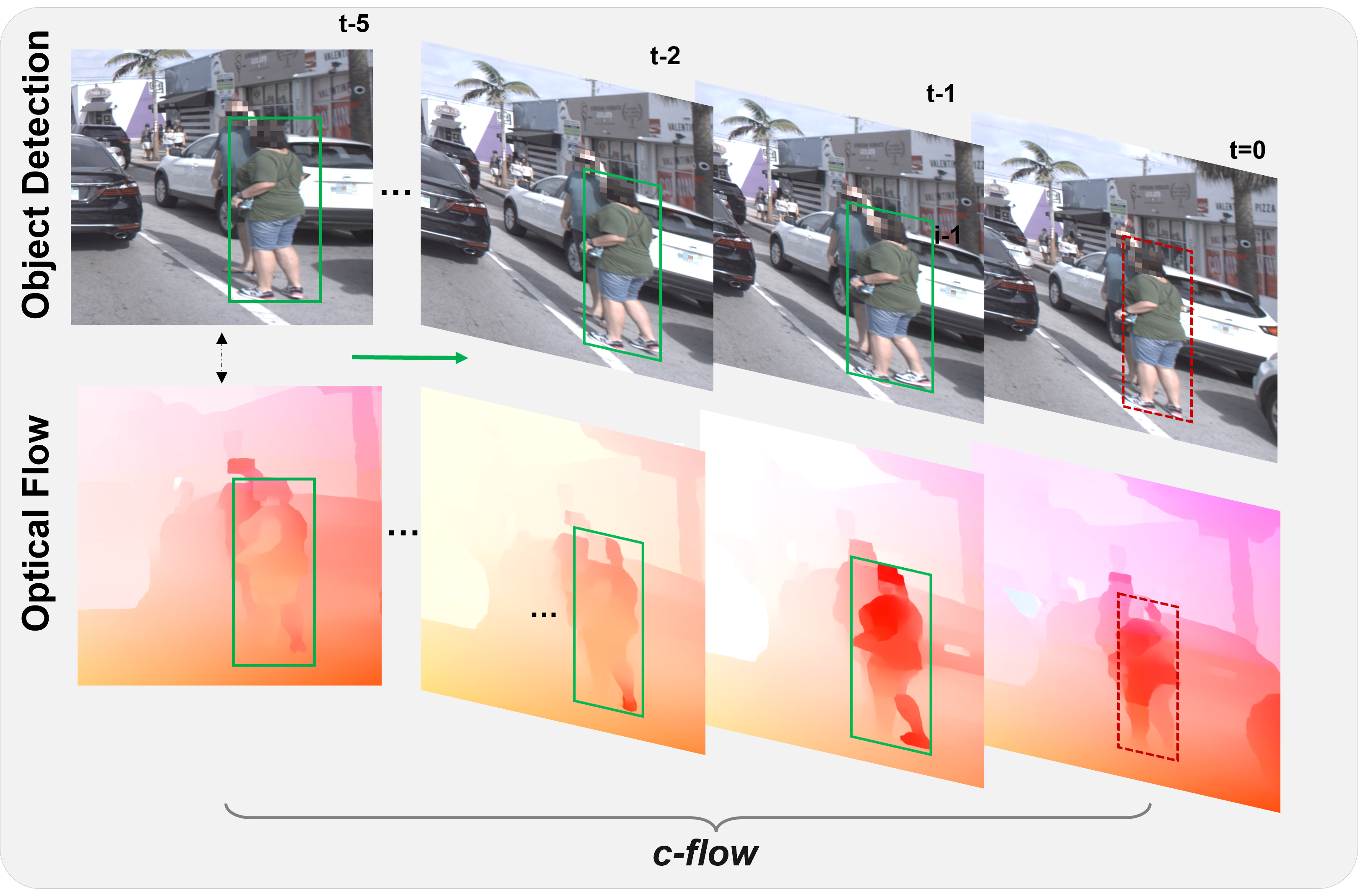}
	\caption{We present a novel metric, called \cflow, to quantify the credibility of pedestrian bounding boxes. Therefore, we exploit temporal information from optical flow (bottom) and a \textcolor{ForestGreen}{series of pedestrian detections} ($\mathfrak{B}_{\textit{pred,t}}$) to rate whether it is credible to have a \textcolor{Burgund}{misdetection} at \textbf{t=0}, \ie~a false negative in our case. Thus, \cflow provides an supplementary signal that helps to uncover prospectively challenging, safety-critical \textcolor{Burgund}{misdetections} (\cflow$\xrightarrow~0$). In the optical flow maps, the intense red color illustrates particularly high, relative motion that we leverage for the metric design.}
	\label{fig:flow}
\end{figure}

However, in the safety-critical domain of AD, failure cases where the system fails to detect objects (so-called \emph{false negatives}) or where the model produces spurious detections (so-called \emph{ghost objects} or \emph{false positives}) during operation continue to occur~\cite{NTSB}.
Hence, both of these erroneous detections not only degrade the overall performance but, in the worst case, lead to a hazardous situation. 
As an example, failing to detect a relevant pedestrian without a warning from the system could result in a safety-critical event such as a collision (as depicted in Fig.~\ref{fig:flow}).
As of now, the development-time evaluation in our previous work~\cite{LGH+22} showed that a state-of-the-art detector tends to produce several misdetections in safety-critical sequences.
This, in particular, could potentially lead to a loss of the pedestrian track at runtime without notice.
For mitigating such cases, our work is motivated by the following research question:

\textit{How can we identify such detection errors in an unsupervised manner and determine whether detections are indeed credible?}

Assessing credibility of detections in an unsupervised manner (\eg~in an AV) implies that we can no longer leverage information from ground truth annotations. 
Instead, we need another, supplementary source of information that can complement the input signal. 
Contemporary works advocate for runtime monitoring of an auxiliary signal to detect potential problems as early as possible, so the AV can adapt his behavior accordingly~\cite{hawkinsrichard,siefke}.
A concrete implementation is proposed by Geissler~\etal~\cite{geissler} where the authors present an approach to monitor input faults of an object detector leveraging knowledge from hidden layers and hardware memory.
Yet, we want to avoid a tight coupling to a specific hardware, and propose a method to verify detections from any object detector.
Hence, in line with works that use optical flow to assess the consistency of detections~\cite{Nishimura2021SDOFTrackerFA} and segmentation masks~\cite{Varghese2020}, we introduce a flow-based method to identify potential errors of our object detector.
As optical flow measures the perceived relative motion of objects between two subsequent images from a sequence, we thereby exploit the time- and space-consistency of objects in the real world and embed it as an auxiliary signal in our proposed method.

An exemplary sequence consisting of RGB images (top) and optical flow maps (bottom) for a safety-critical pedestrian is shown Fig.~\ref{fig:flow}.
Here, the flow maps illustrate the created flow within the predicted bounding box ($\mathfrak{B}_{\textit{pred}}$) due to the pedestrians' motion relative to the AV (intense red color).
Since the quality of optical flow calculation is better in short ranges, as more pixels exist for objects, using the flow to assess the credibility of detections is also a natural choice for our use case where safety-critical pedestrians occur in short ranges from the AV.

In this paper we present a novel metric, called \cflow, for quantifying the credibility of 2D bounding box detections by using information from optical flow.
In particular, we \textit{(i)}~apply \cflow~to a predicted bounding box $\mathfrak{B}_{\textit{pred}}$ and determine whether $\mathfrak{B}_{\textit{pred}}$ is indeed a credible true positive~(TP) detection. 
\textit{(ii)}, to detect cases of false negative~(FN) candidates after a track of successful pedestrian detections,
we propose a technique to infer a 
\textit{hypothesized} bounding box $\mathfrak{B}_{hyp}$ from past predictions.
This allows us to apply \cflow to $\mathfrak{B}_{hyp}$ for assessing the credibility of false negative detections.

We provide an thorough experimental evaluation of \cflow based on the Argoverse 1.1 dataset~\cite{Argoverse} and a state-of-the-art RetinaNet~\cite{retina} for object detection. 
Accordingly, we \textit{(i)}~demonstrate the validity of our novel metric in a supervised manner where ground truth information is accessible.
Our results show that \cflow can successfully discriminate between true positives and safety-critical false negatives for most pedestrian samples in vicinity of the AV.
For ambiguous cases of true positives by means of \cflow, a qualitative analysis shows how the metric is eligible to identify cases with an unreasonable labeling policy (\eg~distant or heavily occluded pedestrians), \ie~aiding as a tool to provide dataset insights.
\textit{(ii)}, we evaluate the metric in scope of prospective runtime applications.
Hence, as ground truth information is unavailable, we demonstrate how the approach with $\mathfrak{B}_{hyp}$ is effectively applicable to an unsupervised setting. 


\section{Methodology}
\label{sec:framework}
In this section, we present our novel credibility metric. \cflow leverages changes in optical flow to quantify the credibility of bounding boxes within safety-critical sequences by means of the temporal distance (time-to-collision) between the AV and the respective pedestrian. 
We motivate our methodology for the \cflow metric design in Sec.~\ref{sec:motivation}, followed by implementation details in Sec.~\ref{sec:metricdesign}.
In Sec.~\ref{sec:bbox}, we illustrate the approach to compute so-called \textit{hypothesized} bounding boxes $\mathfrak{B}_{hyp}$ for unsupervised use cases.

\subsection{Motivation on \textit{c-flow}}
\label{sec:motivation}
\begin{figure}[t!]
	\centering
	\includegraphics[width=0.7\columnwidth]{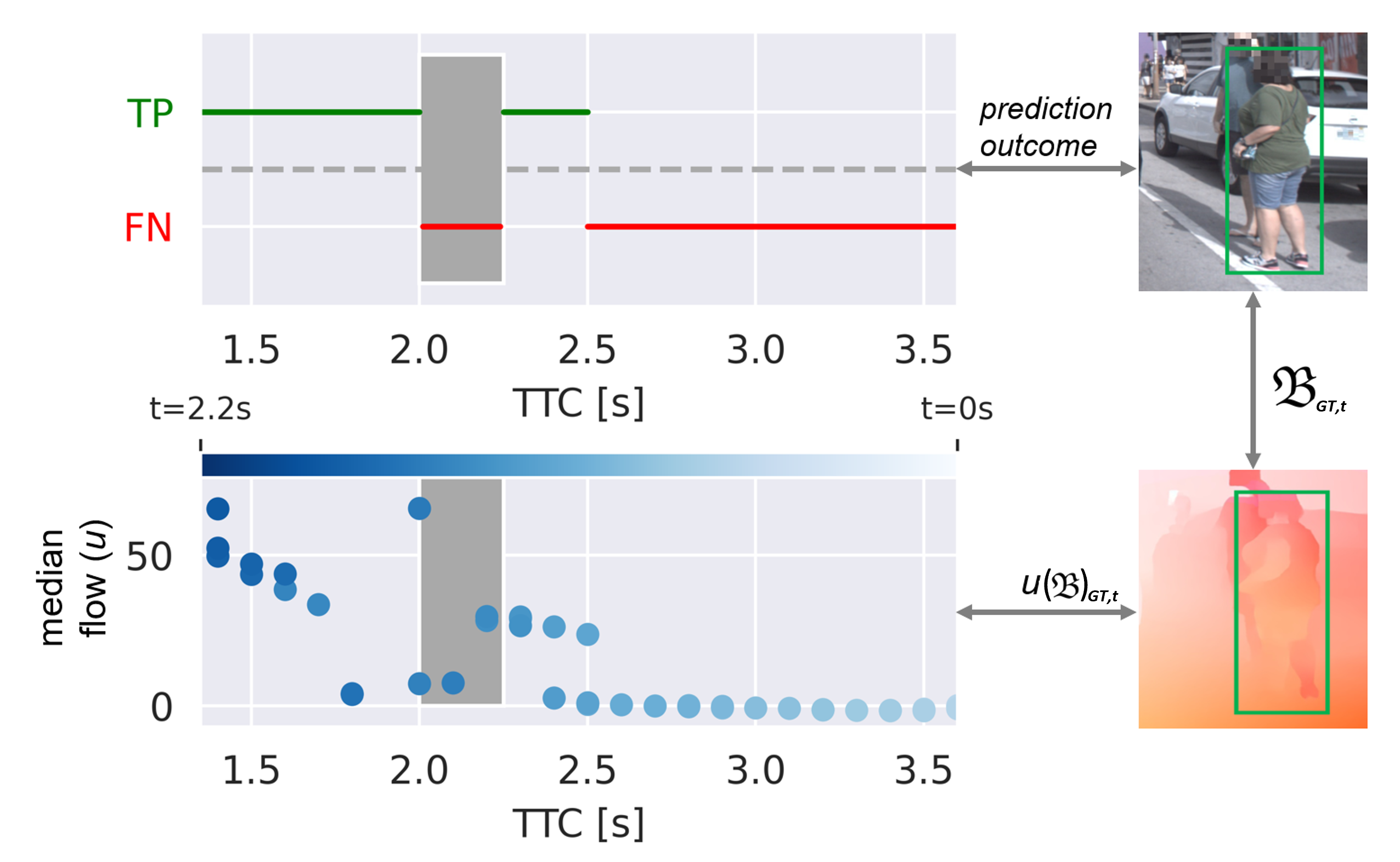}
	\caption{Selected pedestrian track (red star sequence from Fig.~\ref{fig:crit}) showing evolution over time (in color) for a pedestrian bounding $\mathfrak{B}_\textit{GT,t}$ illustrating \textit{(i)} the classification outcome (top) and \textit{(ii)} the median score $u$ gathered from the optical flow map at $\mathfrak{B}_\textit{GT,t}$. The gray window highlights sudden fluctuations in $u$ that correlate with the switch between classification outcomes at, \eg, TTC$\approx$2.25$s$.}
	\label{fig:track}
\end{figure}

Let us provide a motivating example for using flow to assess the credibility of pedestrian detections.
To this end, we select one safety-critical pedestrian track as shown in Fig.~\ref{fig:track}~(for details see Sec.~\ref{sec:sequences} and Fig.~\ref{fig:crit}), with time indicated by the colormap.
The respective criticality as time-to-collision (TTC) is on the x-axis. 
As this is a real pedestrian track with associated ground truth (GT), the classification outcome that is shown in the top plot for a bounding box $\mathfrak{B}_\textit{GT,t}$ is at different times $t$ either a TP or a FN (\ie~not detected).

The upper plot depicts the typical behavior of a longer track: No detections for the pedestrian at larger distances (\ie~high TTC) with the first successful detection at TTC$=2.5s$ after the AV has approached the pedestrian\footnote{Please note, since near range pedestrians have a better quality of optical flow due to a higher pixel count, the optimal use of optical flow is aligned with our motivation to perform a flow-based credibility assessment on $\mathfrak{B}_\textit{GT}$ for pedestrians close to the AV.
}.
In the lower plot, we use a pedestrian bounding box $\mathfrak{B}_\textit{GT,t}$ from our selected track at time $t$ in $t\in [1.25\,s,3.5\,s]$ to construct the corresponding data points. 
Therefore, we apply $\mathfrak{B}_\textit{GT,t}$ to the optical flow map to define the area of interest where we calculate the optical flows' median score $u$\footnote{For further discussion, we only utilize the horizontal flow map, as the vertical flow showed only oscillating displacements between consecutive images, \ie~no meaningful vertical shift in $\mathfrak{B}_\textit{GT}$.}.

On the one hand, in the temporal plot around the highlighted gray window, we denote significant changes in $u$, which are a strong indicator for errors concerning the consistency of objects between images~\cite{Varghese2020}.
On the other hand, if we look at the detection type, we can derive that the rapid changes in $u$ correlate with the classification outcome switch from TP to FN at TTC$\approx2.25\,$.
As optical flow estimates the relative motion of the pedestrian between consecutive images, sudden changes in $u$ may refer to abrupt variations in the perceived motion. This may indicate unexpected changes in the scene such as partial or complete vanishing of the pedestrian due to, \eg~a sudden occlusion.
This conveys a difficult case for our detector-under-test as occlusions are a well-known challenge for object detectors.

In summary, this example shows that optical flow can provide a supplementary signal for analyzing the credibility of predictions: Small changes in $u$ indicate a continuous existence of a pedestrian detection over images, whereas sudden fluctuations in $u$ may help to identify generally difficult cases for the object detector.

Consequently, we want to discriminate correct TP from prospective failure cases where it is non-credible to have a missing detection.
This motivates our work to exploit the sudden changes in optical flow to design a metric, which we call \cflow in the following, that can be applied to a bounding box to assess the credibility.
With the novel metric we aim to achieve the discrimination between TP detections (high \cflow score) and non-detections, \ie~FN (low \cflow score).


\subsection{Metric Design: \textit{c-flow}}
\label{sec:metricdesign}
Based on this underlying idea, we design the \cflow metric for individual pedestrians and the corresponding bounding box.
Therefore, our metric design is guided by the following desiderata: Since we want to assess a credibility score and thus, to discriminate successful detections from cases of FN candidates, we design our novel metric to be \cflow$\xrightarrow~0$ for FN and \cflow$\xrightarrow~1$ for TP, respectively.

In the first step, we quantify the continuity of pedestrian detections among consecutive images.
Therefore, we utilize the given track information from Argoverse 1.1 (see Sec.~\ref{sec:datasetsdet}), \ie~we leverage $\mathfrak{B}_\textit{GT}$ from GT for missing detections (FN).
Concretely, we consider a pedestrian track of a certain length with pairs of \textit{(i)} the pedestrians' bounding box $\mathfrak{B}_\textit{GT}$ and \textit{(ii)} calculated optical flow median scores $u$ (extracted from flow maps within $\mathfrak{B}_\textit{GT}$).
To include the sudden flow changes in $u$ leveraging a series of successful detections across time, we construct our metric on the basis of a time window $w$, that encompasses the past $k$ images with $w=[t_{k},t_{k-1}...,t_{1},t_{0}]$, \ie~$\mathfrak{B}_\textit{GT,t}$ for each image at $t$ in $w$.

\begin{figure}[t!]
	\centering
	\includegraphics[width=0.5\columnwidth]{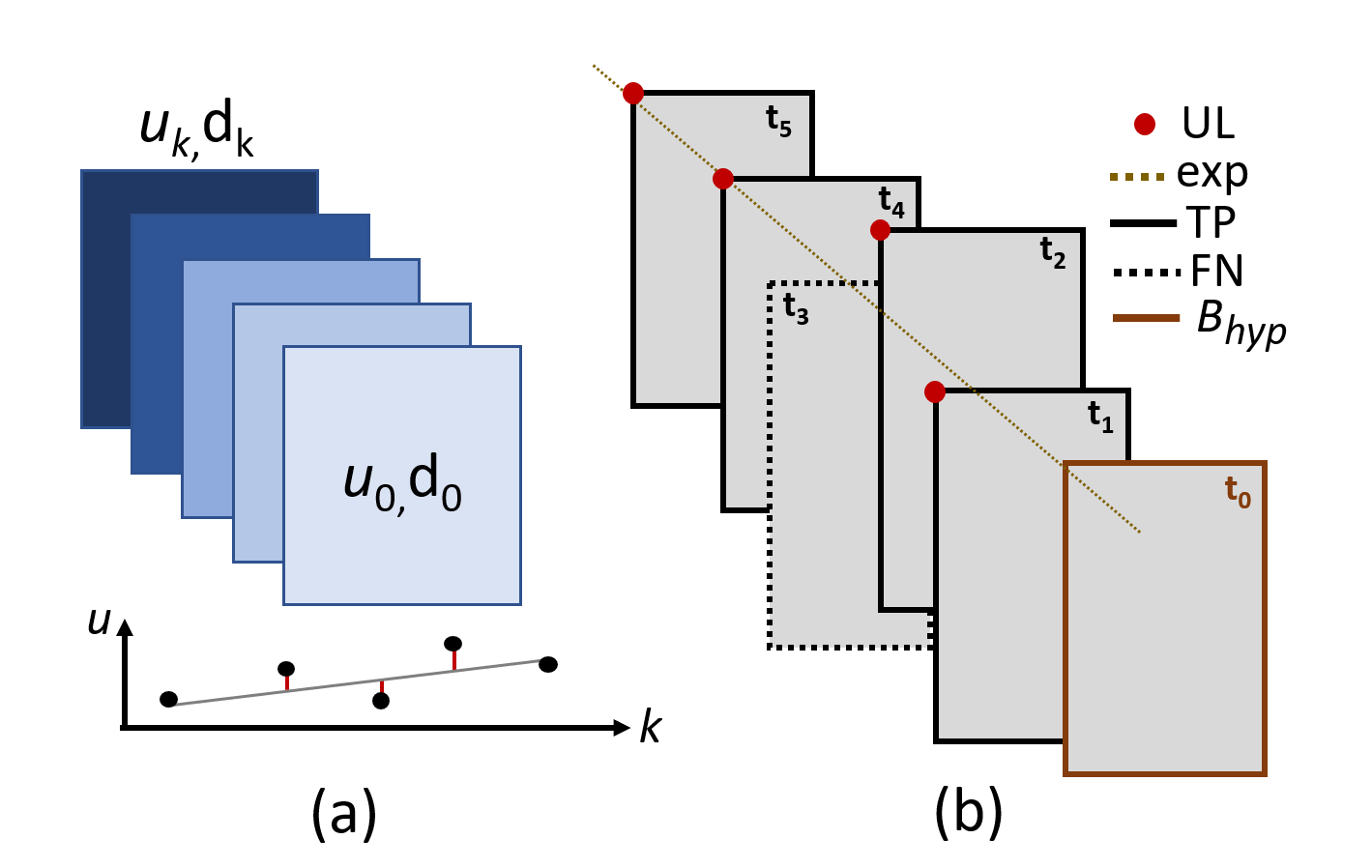}
	\caption{\textit{(a)} Leveraging linear regression to determine the variability in $u$ to construct \cflow. \textit{(b)} For missing detections at $t_0$, we apply the methodology of hypothesized bounding boxes ($\mathfrak{B}_{hyp}$) to infer the required $\mathfrak{B}_{hyp}$. We extract the upper left corner (UL) of past predictions to extrapolate the hypothesized UL position at $t_0$.}
	\label{fig:halluc}
\end{figure}
In the second step, we want to quantify the flow change across $w$ up to the current image at $t_0$.
Therefore, we leverage the calculated $u$ for each $\mathfrak{B}_\textit{GT,t}$, to determine the variability of the flow in $[u_{k},u_{k-1}...,u_{1},u_{0}]$ over $w$.
Given the underlying assumption of linear progression of the movement across $w$, \ie~a linear change of flow in $w$, we now compute a linear regression on  $[u_{k},u_{k-1}...,u_{1},u_{0}]$ as illustrated in Fig.~\ref{fig:halluc}~\textit{(a)}. 
After applying the linear regression, we estimate the residuals $r=[r_{k},r_{k-1}...,r_{1},r_{0}]$ (error bars in red) between regressed $u$ scores and $[u_{k},u_{k-1}...,u_{1},u_{0}]$ (black dots), respectively, to quantify the deviation from the linear progression, \ie~a sudden flow change that indicates a difficult case.
To define the overall error $\epsilon$ over $w$, we calculate $\epsilon=\sum_{i=0} ^{k} r_{i}$.

In the previous steps, we have identified a strong correlation between rapid changes in $u$ and switches in the classification outcome.
However, as flow changes over $w$ also correlate with the change in size of $\mathfrak{B}_\textit{GT,t}$ (\ie~a larger bounding box captures more pixels to estimate $u$), we must consider this change in box size.
Therefore, we inject supplementary information on $\mathfrak{B}_\textit{GT,k}$ size into the metric design.
To describe the change in size, we utilize the diagonals between the two most recent bounding boxes $d_0$ and $d_1$ (at $t_0$ and $t_1$).
Thus, we introduce $\Delta{d}_{t_{0,1}}={d}_0-{d}_1$ to measure the difference.

Based on the desiderata, $\Delta{d}_{t_{0,1}}$, and $\epsilon$, we can now show the complete formula for the credibility metric in Eq.~\ref{eq:equation}.
Please note, as we bring $\Delta{d}_{t_{0,1}}$ and $\epsilon$ in relation to each other, we first normalize both parameters.
Finally, we propose the sigmoid function 
\begin{equation}
\cflow(\Delta{d}_{t_{0,1}},\epsilon) =\text{sigmoid}(\frac{\Delta{d}_{t_{0,1}}}{\epsilon})
\label{eq:equation}
\end{equation}
that introduces a non-linear, gradual transition from 0 to 1, normalizing \cflow to the unit interval.

\subsection{Handling the Absence of Ground Truth in Unsupervised Use Cases}
\label{sec:bbox} 
Up to now, we have used GT bounding boxes for analyzing the FN cases from Fig.~\ref{fig:track}.
Thus, we have applied the \cflow metric using a window of GT bounding boxes, to assess the credibility that a predicted bounding box $\mathfrak{B}_\textit{pred}$ is missing at $t_0$.
However, let us assume that we do not have an available $\mathfrak{B}_\textit{pred}$ for the current image at $t_0$, as, \eg, when the detector is running in a vehicle or when performing an unsupervised data analysis.
Thus, for the current track-under-test for a particular pedestrian, we want to assess whether the lack of $\mathfrak{B}_\textit{pred}$ at $t_0$ is credible or whether it is a prospective FN.
Therefore, to compensate the absence of $\mathfrak{B}_\textit{pred}$ in the \cflow computation, we introduce the concept of \textit{hypothesized} bounding boxes, \ie~$\mathfrak{B}_{hyp}$, that are extrapolated from available, past $\mathfrak{B}_\textit{pred}$ for extracting the optical flow crops.
In Fig.~\ref{fig:halluc}~\textit{(b)}, we can see a sequence of past, predicted $\mathfrak{B}_\textit{pred}$ at $t_5$, $t_4$, $t_2$, and $t_1$. 

As a first step, we extract the upper left corner (UL) of previous TP detections. 
We focus on UL as it is the most probable, visible corner for cases with horizontal or vertical possible occlusions in the front camera perspective~\eg, a pedestrian standing behind a car on the right side of the street.
Next, we perform an extrapolation (see~Fig.\ref{fig:halluc}~\textit{(b)}) of the pixel coordinates to hypothesize UL for $t_0$. 
Please note, we consider the previous five ($k=5$) frames using the detection window $w$ from Sec.~\ref{sec:metricdesign}. However, the algorithm does not require all $\mathfrak{B}_\textit{pred}$ in $w$ to exist (see skipped detection at $t_3$). 
It necessitates only a minimum of two existing $\mathfrak{B}_\textit{pred}$ in the detection window (num. TP $ \geq 2$) to fit the extrapolation line \textit{exp}.
Thus, this approach would also work after the first \textit{two} successful detections after the pedestrian track was initially opened.

To hypothesize UL at $t_0$, we need to estimate the positional shift from the last available $\mathfrak{B}_\textit{pred}$ ($t_1$ in our case). 
Therefore, to determine the mean positional shift between $n$ consecutive detections, we calculate the mean pixel displacement, \ie~distance between the first ($t_5$) and the last available $\mathfrak{B}_\textit{pred}$ ($t_1$) from our example averaged over $n=4$.  
Finally, we apply the determined shift along \textit{exp} to hypothesize UL at $t_0$.
For the initial proof-of-concept, we assume a small change in dimension of $\mathfrak{B}_{hyp}$ to $\mathfrak{B}_\textit{pred}$ from the previous image at $t_1$ and set the width and height of $\mathfrak{B}_{hyp}$ to be $w_{t_1}$ and $h_{t_1}$, respectively.
%
As a last step, we extract the estimated crop to calculate $\cflow(t_0)$.
Please note, in Fig.~\ref{fig:halluc}~\textit{(b)}, $t_3$ could have been replaced by $\mathfrak{B}_{hyp}$ to boost accuracy for runtime applications.
Further approaches to regress the dimensions of $\mathfrak{B}_{hyp}$ from $w$ we leave for further work.

\begin{figure}[t!]
	\centering
	\includegraphics[height=0.3\columnwidth]{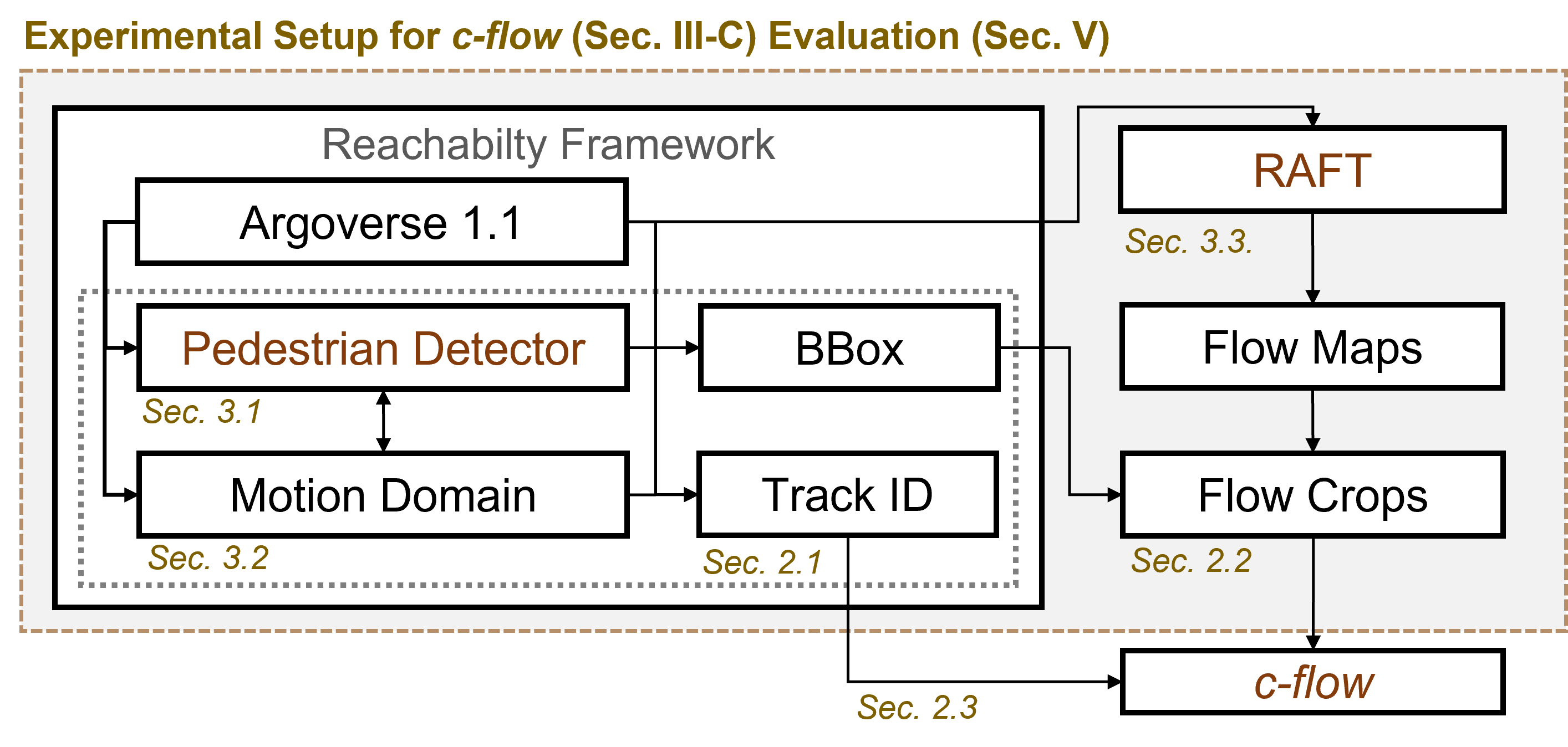}
	\caption{Experimental setup for \cflow~evaluation. We extend our reachability framework (RF) from~\cite{LGH+22} by a DNN-based optical flow estimation using RAFT. We perform our evaluation on the basis of identified, potentially safety-critical pedestrian tracks (Track ID) with respect to TTC derived from the motion domain.}
	\label{fig:framework}
\end{figure}

\section{Experimental Setup}
\label{sec:experimentalsetup}
In the following, we leverage the setup from Fig.~\ref{fig:framework} to evaluate our novel \cflow metric. In our setup we utilize \textit{(i)} the nuImages dataset to pre-train our RetinaNet from Sec.~\ref{sec:datasetsdet}, \textit{(ii)} an adapted sequence selection approach using our reachability framework (RF)~\cite{LGH+22} in Sec.~\ref{sec:sequences}, and \textit{(iii)} the incorporation of RAFT from Sec.~\ref{sec:flowestimation} to generate optical flow maps.

\subsection{Datasets and the Pedestrian Detector}
\label{sec:datasetsdet}
For the training of our pedestrian detector we use nuImages~\cite{nuscenes2019}\footnote{\url{https://github.com/nutonomy/nuscenes-devkit}}.
Thereby, we do not train on the Argoverse 1.1 dataset and thus, we can employ all splits from Argoverse 1.1~\cite{Argoverse}\footnote{\url{https://github.com/argoverse/argoverse-api}} with Argoverse HD 2D annotations~\cite{argoverseHD} to evaluate \cflow. 
Please note, in our evaluation, we only use images from the \texttt{ring\_front\_center} camera.
Afterwards, we estimate the matching LiDAR GT (different frequency at 10 Hz) using \texttt{image\_list\_sync} to extract physical properties of the AV and the pedestrians, \ie~information that is required to calculate the TTC.
We implement the RetinaNet\footnote{\url{https://github.com/yhenon/pytorch-retinanet}} using PyTorch and employ the following training protocol: We utilize the ResNet-50 backbone, the Adam optimizer and the applied learning rate of $1e^{-5}$, the \texttt{reduceOnPlateau} scheduler (patience=2) from the \texttt{optim.module}, and we train RetinaNet for 200 epochs.
The trained model achieved a reasonable performance of 0.31 $AP50$ for the pedestrian class on the respective nuImages validation split.
For the pedestrian class in Argoverse 1.1, the model reached a performance of 0.35 $AP50$ and 0.34 $AP50$ on the train and validation split, respectively.
Please note that we do not strive for highest performance as we are interested in a variety of cases for our \cflow metric.

\subsection{Pedestrian Track Selection on Argoverse 1.1}
\label{sec:sequences}
As we want to evaluate \cflow on individual pedestrian tracks of different criticality, we first need to extract the respective sequences~\wrt~TTC.
Therefore, we leverage our RF as described in previous work~\cite{LGH+22} to calculate reachable sets and its intersection (between AV and pedestrians) utilizing physical properties as, \eg, velocity and position and semantic map information with explicit lanelets. 

\begin{figure}[t!]
	\centering
	\includegraphics[width=.65\columnwidth]{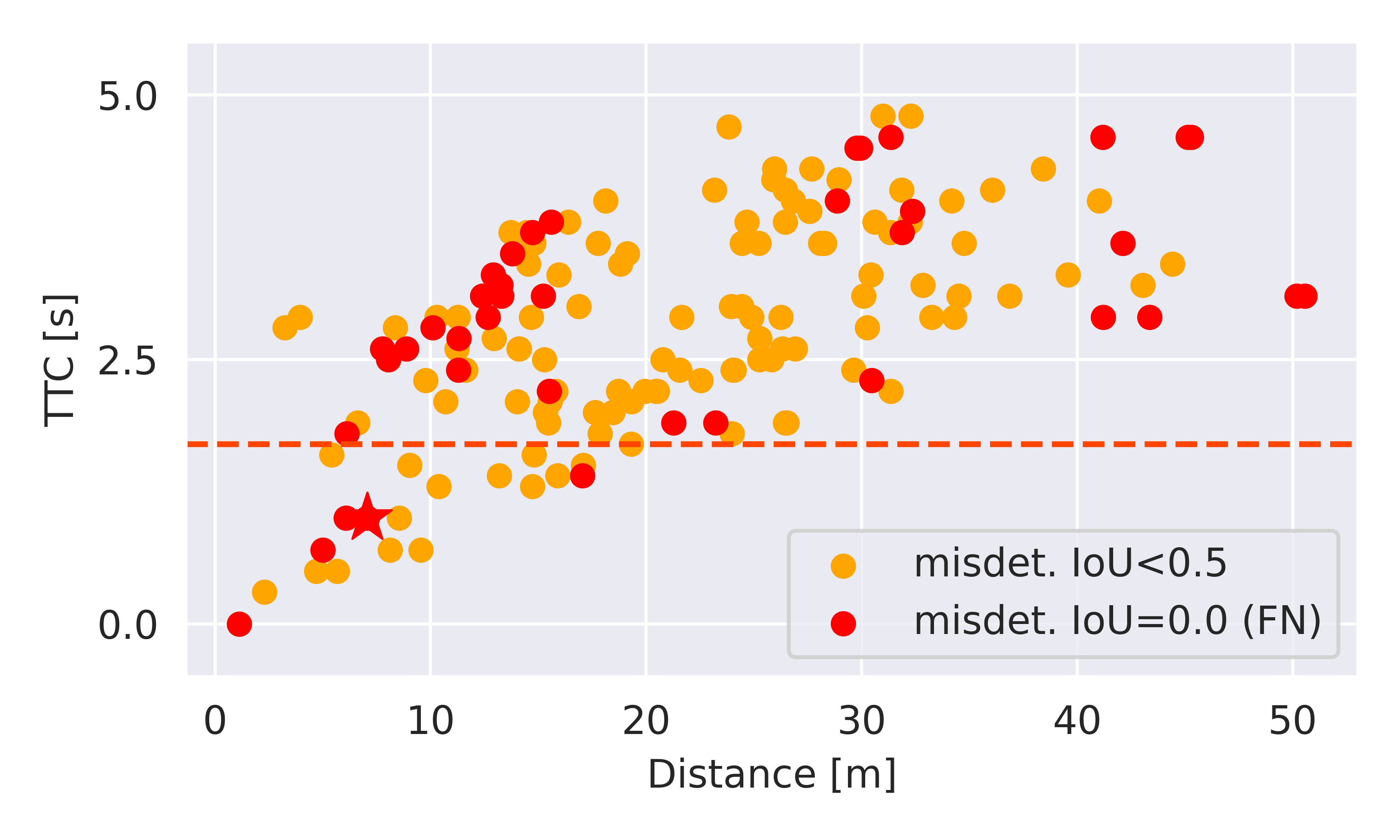}
	\caption{Extracted pedestrian tracks from Argoverse 1.1 using the reachability framework. Each data point represents an interaction defined by the most critical pedestrian misdetection~\wrt~TTC and distance. The dotted line separates critical (TTC$<2\,s$) vs. non-critical interactions. Misdetections with a poor detection quality (0$<$IoU$<$0.5) are highlighted in orange and FN (IoU=0) are marked in red.} 
	\label{fig:crit}
\end{figure}

Fig.~\ref{fig:crit} shows the total sequence count that encompasses 142 of such tracks with 17 highlighted interactions that include critical scenarios, \ie~completely missed pedestrians (red) or insufficient performance with 0$<$IoU$<$0.5 (orange).
Note that we introduce an improved approach that increases the sequence count from 32~(using~\cite{LGH+22}) to 142 available sequences.
The novel implementation incorporates a matching refinement to find corresponding tracks between Argoverse 1.1 and Argoverse HD.
For instance, Argoverse HD creates a new ID as soon as the tracked pedestrian reappears after a full occlusion. 
Thus, multiple tracks in Argoverse-HD may describe sub-parts from a single Argoverse 1.1 sequence. 
Our adaptation alleviates this inconsistency and summarizes multiple tracks from Argoverse-HD to one unique sequence that is matched with its counterpart from Argoverse 1.1.

\subsection{RAFT: Optical Flow-based Motion Estimation}
\label{sec:flowestimation}
We employ the RAFT~\cite{Teed2020RAFTRA}\footnote{\url{https://github.com/princeton-vl/RAFT}} model to determine the optical flow on the basis of two consecutive images.
The model was already trained on KITTI\footnote{\url{https://www.cvlibs.net/datasets/kitti/}} with a performance of 7.51 $epe$~(endpoint error) and a $F1$ score of 0.269. 
Thereafter, for each pedestrian sequence from~\ref{sec:sequences}, we infer image pairs to calculate a flow map as shown in Fig.~\ref{fig:flow} (bottom).
For visualization purposes, we use \texttt{flow\_uv\_to\_colors}~\cite{Teed2020RAFTRA}.

\section{Experimental Results}
\label{sec:results}
In this section, we conduct an evaluation of the \cflow~metric.
In Sec.~\ref{sec:offline}, we utilize the safety-critical pedestrian tracks to perform a thorough validation of the metric in supervised manner where detailed GT information is available.
In Sec.~\ref{sec:online}, we evaluate whether \cflow can provide accurate information for FN detections in an unsupervised setting when using  $\mathfrak{B}_{hyp}$ instead of actual GT information.

\subsection{\textit{c-flow} Evaluation: TP vs. FN}
\label{sec:offline}
We start by computing \cflow values for all pedestrians in the selected pedestrian tracks of different criticality using the predicted $\mathfrak{B}_\textit{pred}$ for TP, and GT annotations for FN.
As discussed in Sec.~\ref{sec:metricdesign}, the goal is to detect cases of FN candidates after a series of successful detections (TP) by means of a low \cflow, \ie~\cflow$\xrightarrow~0$.
At the same time, cases of TP should have a high \cflow$\xrightarrow~1$. 
Ideally, depending on a set threshold $\xi$, \cflow would be able to discriminate between \textit{(i)} successful detections with $\cflow > \xi$ and \textit{(ii)} cases of FN candidates with \cflow $\leq \xi$.

\begin{figure}[t!]
	\centering
	\includegraphics[width=0.65\columnwidth]{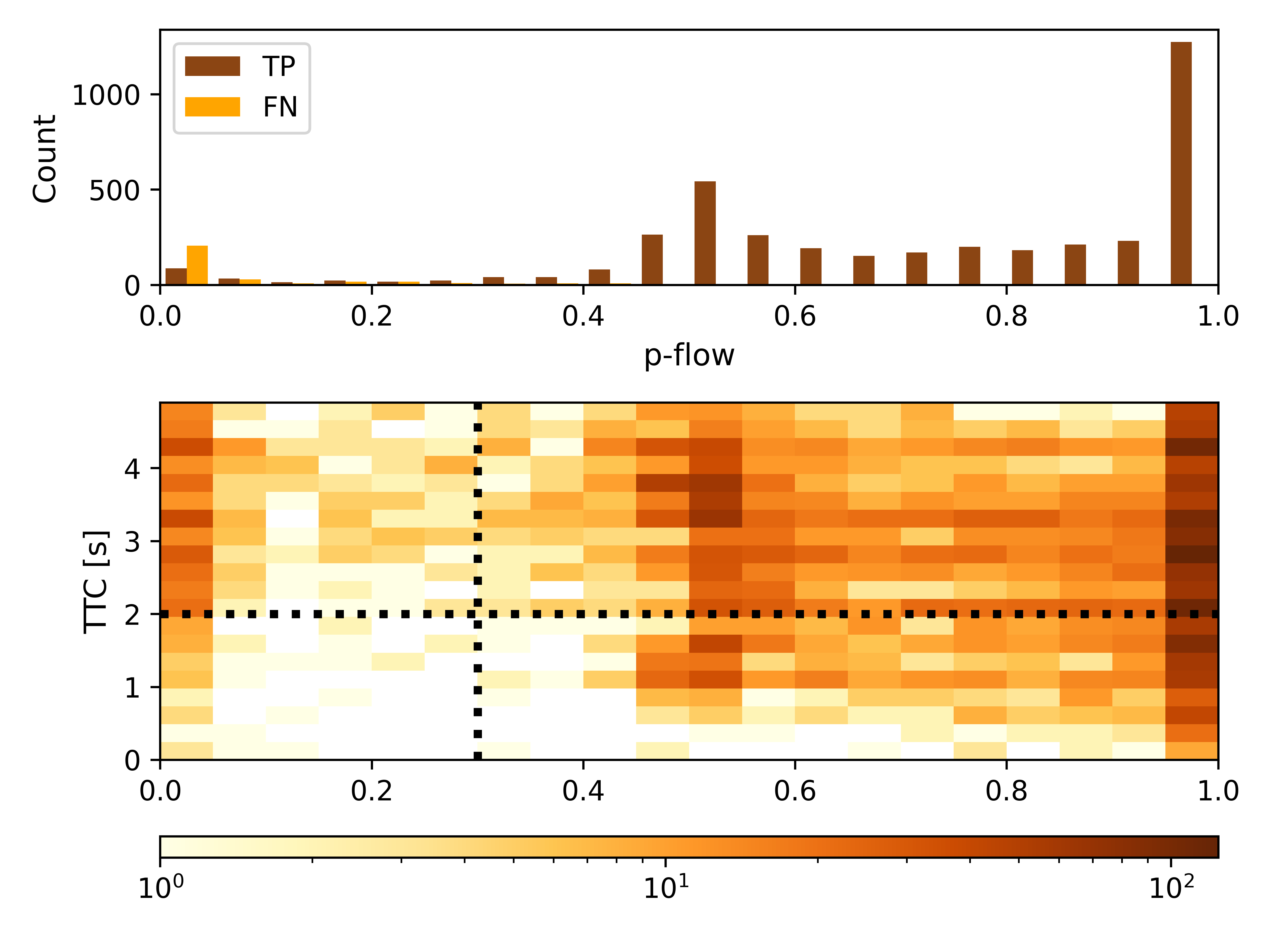}
	\caption{Histogram (top) shows the distribution of the \cflow scores conditioned on the prediction types (FN or TP). The heatmap (bottom) depicts the \cflow scores with respect to their TTC criticality. The dotted lines divide the heatmap into regions of different criticality, constrained by the TTC = 2$s$.}
	\label{fig:barplot_TP_FN}
\end{figure}
As a first step, we evaluate the \cflow scores using Fig.~\ref{fig:barplot_TP_FN}.
The histogram (top) shows the distribution of the \cflow scores conditioned on classification GT (TP or FN).
In the histogram, we can see that all FN samples received a \cflow~$<\,\xi= 0.4$ and that almost all TP have a \cflow score $\geq \xi=0.5$. 
However, in comparison to the occurrence of FN with a \cflow mostly in the interval~$\in [0.0, 0.3]$, we can also see a spread of TP over lower \cflow scores. 

Let us now investigate, how the selection of a concrete threshold on \cflow affects the identification of FN in safety-critical pedestrian tracks \wrt~TTC.
The heatmap on the bottom of Fig.~\ref{fig:barplot_TP_FN} visualizes the amount of \cflow scores (x-axis) in relation to the TTC value (y-axis). 
The figure also shows the critical zone by the TTC threshold of 2$\,s$ (horizontally dotted line) and a \cflow threshold of $\xi=0.3$ (vertically dotted line). 
From the heatmap, we can derive that for high criticality (TTC $\leq\,$ 2$\,s$), the \cflow value is either close to zero or shows \cflow scores  $\geq \xi= 0.5$.
Only for larger TTC values, we can see that the \cflow scores are less discriminating. 

\begin{figure}[t!]
	\centering
	\includegraphics[width=0.7\columnwidth]{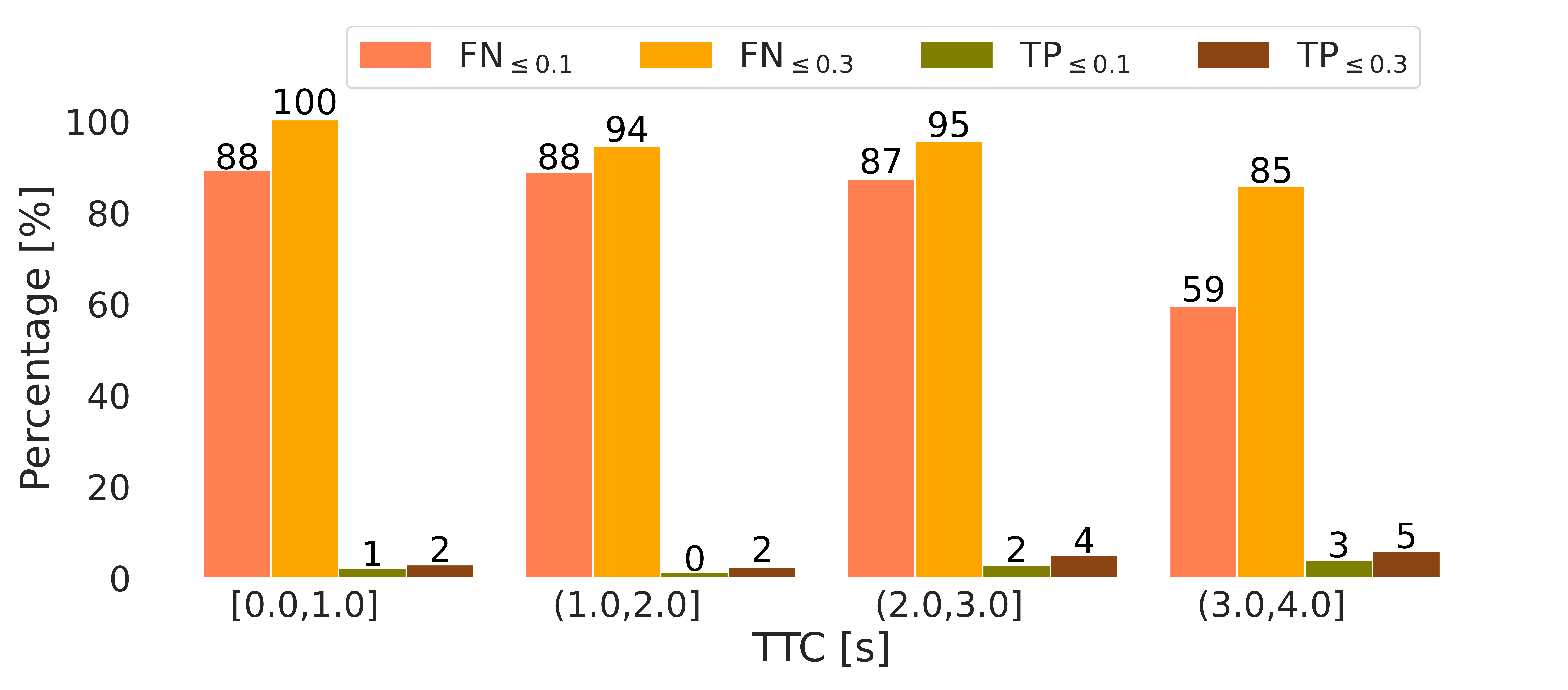}
	\caption{Bar plot depicts the percentage of identified prediction types (TP and FN, out of annotated GT) among TTC intervals of different criticality for two \cflow~thresholds: $\xi=0.1$ and $\xi=0.3$.}
	\label{fig:rel_p_TTC}
\end{figure}

As a second step, in Fig.~\ref{fig:rel_p_TTC}, we study the relative distribution of identified TP and FN samples (y-axis) with respect to TTC intervals of different criticality (x-axis). 
Therefore, we select two concrete thresholds $\xi_1$ and $\xi_2$ to study how the selected threshold affects the trade-off between correctly identified TP and FN on the basis of \cflow.
From the result in Fig.~\ref{fig:barplot_TP_FN}, we are particularly interested in: \textit{(i)} $\xi_1=0.1$ as it defines the interval with almost all FN samples by means of a \cflow$\xrightarrow~0$ but also TP occurrences and \textit{(ii)} $\xi_2=0.3$ as for $0.1<\cflow<0.3$ there is yet a small number of TP and FN cases.

Now, we apply $\xi_1=0.1$ and $\xi_2=0.3$, respectively, to investigate the impact on the assigned classification outcome among specified TTC intervals.
Therefore, we focus on the two most critical intervals (TTC$\leq1\,s$ and $1\,s<\text{TTC}\leq~2\,s$), \ie~the first two sets of bars on the left, for the two most critical TTC intervals. 
We can see that with $\xi_1$, 88\% of all FN are correctly identified by means of a low \cflow, whereas 1\% and 0\% of TP samples are falsely marked as non-credible, respectively.
Raising the \cflow threshold to $\xi_2$ results in 100\% and 94\%, \ie~all and almost all FN are marked as non-credible with a low \cflow
for the two most critical TTC intervals.
However, this comes at the cost of also more TP with a low \cflow. 
For safety-critical zones we can see a maximum of 2\% for this threshold, which is still low overall, yet a four-fold increase in the critical region compared to the threshold of $\xi_1$.

\begin{figure}[t!]
	\centering
	\includegraphics[width=0.65\columnwidth]{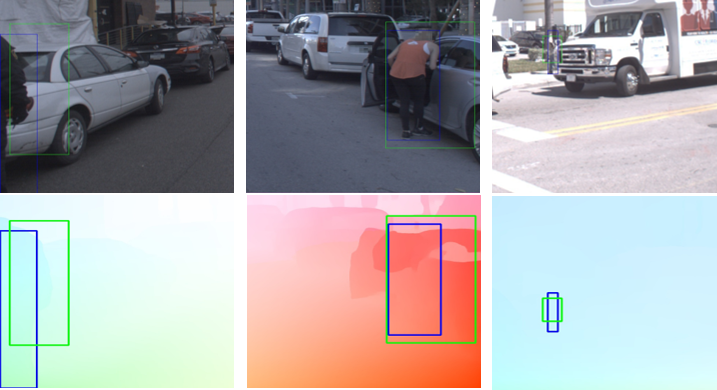}
	\caption{TP samples with a low \cflow score ($<$0.3), to illustrate possible causes for misleading \cflow estimates. Here, we can see pairs of RGB (top) and optical flow (bottom) with predicted bounding boxes (green) and annotations (blue), showing problematic cases of TP including inaccurate predictions and no perceived flow.}
	\label{fig:TP_low_plaus}
\end{figure}

To investigate the underlying reason behind problematic TP with a low \cflow, we analyzed TP detections by means of \cflow$<$0.3 for 35 test images that we gathered from different tracks for variability.
The qualitative results illustrate \textit{(i)} 9 cases of occlusion, \textit{(ii)} 12 cases of distant pedestrians (\ie, minimally perceived flow), and \textit{(iii)} 14 cases of inaccurate detections.
As illustrated in Fig.~\ref{fig:TP_low_plaus} (left and right), multiple effects as, \eg~inaccurate predictions and minimal perceived flow (visualized in blue) may occur simultaneously.  
Thus, artifacts such as occlusion and truncation, no or minimally perceived flow for distant pedestrians, and inaccurate detections (low IoU) may produce misleading estimates due to \eg, poor flow aggregation in $\mathfrak{B}_\textit{pred}$, \ie~TP with low \cflow. 
Thereby, we can see that low \cflow scores for TP may help to identify difficult cases for the object detector during runtime applications in an AV, especially \wrt~to safety-critical pedestrians (left and mid). 
We leave elaborate analyses on concrete corner cases for future work.

\subsection{FN in an Unsupervised Setting}
\label{sec:online} 
Further, we evaluate \cflow's ability to determine FN candidates in pedestrian tracks in an unsupervised setting where no GT information is available.
As described in Sec.~\ref{sec:bbox}, we exploit existing $\mathfrak{B}_\textit{pred}$ for the pedestrian from previous images for estimating a $\mathfrak{B}_{hyp}$ that we use for \cflow.
In this experiment, we compare the resulting \cflow values for $\mathfrak{B}_{hyp}$ to the \cflow values obtained based on ground truth $\mathfrak{B}_{GT}$ to validate that the use of $\mathfrak{B}_{hyp}$ indeed works as intended.
For differentiating the different values in the following, we use $\cflow_{h}$ to denote a \cflow obtained based on $\mathfrak{B}_{hyp}$ and $\cflow_{GT}$ to denote a \cflow obtained based on GT.
\begin{figure}[t!]
	\centering
	\includegraphics[width=0.6\columnwidth]{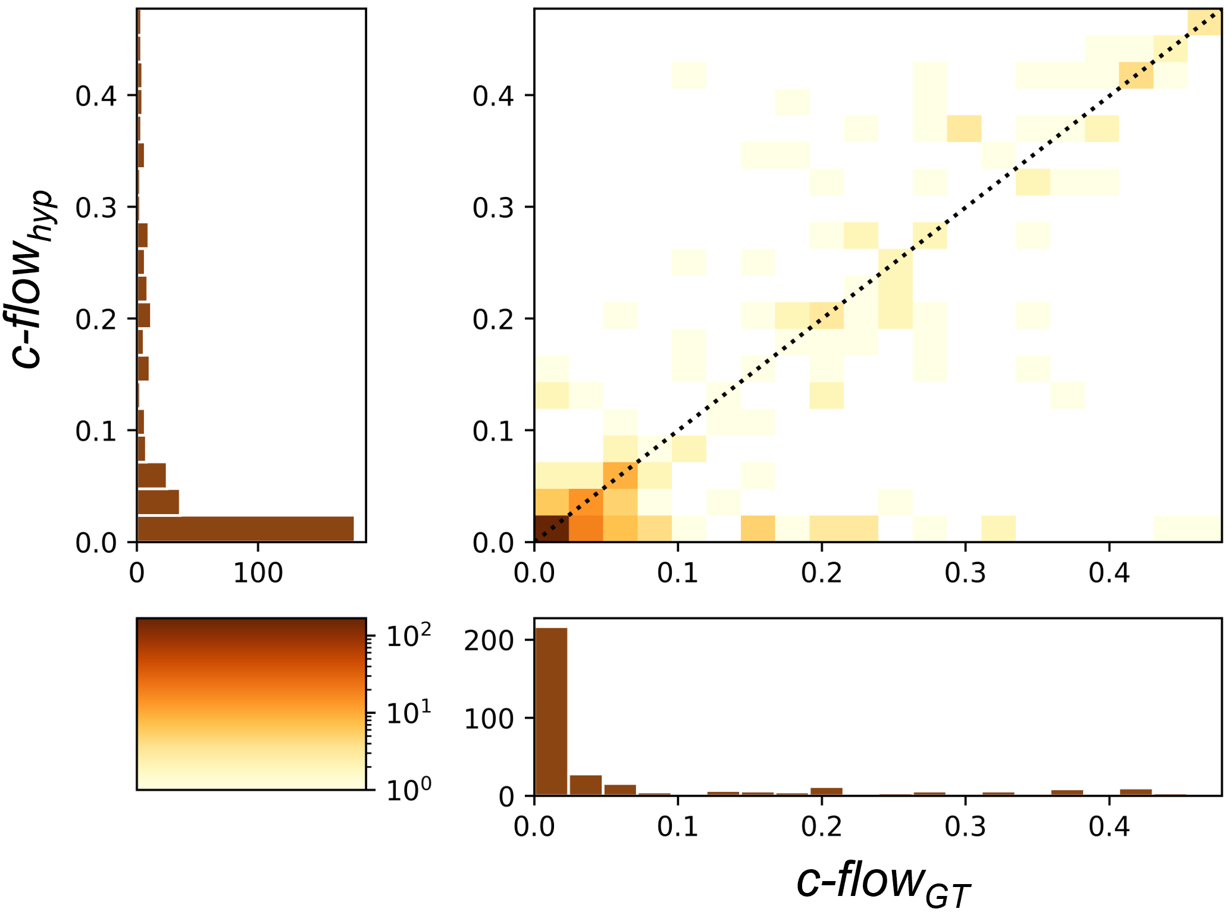}
	\caption{The heatmap indicates the correlation ($\rho=0.83$) of hypothesized $\cflow_{hyp}$ and annotation-based $\cflow_{GT}$, dependent on the count of FN samples. The bar plot on the left and at the bottom, show the distribution of FN samples with respect to $\cflow_{hyp}$ and  $\cflow_{GT}$, respectively.}
	\label{fig:online_eval}
\end{figure}
The heatmap in Fig.~\ref{fig:online_eval} depicts the count of occurring FN samples conditioned on $\cflow_{GT}$ and $\cflow_{hyp}$. 
Ideally, most of the samples should lie on the bisecting line that represents a perfect linear correlation between $\cflow_{hyp}$ and  $\cflow_{GT}$. 
In the heatmap plot, we can see a high correlation (pearson's coefficient $\rho\,=\,0.83$) between $\cflow_{GT}$ and $\cflow_{hyp}$.
Especially for $\cflow\leq \xi_1$, the bar plot illustrates that the majority of FN samples fall within that interval, \ie~we have a high FN sample count, \wrt~$\cflow_{hyp}$ (in the top left) and $\cflow_{GT}$ (at the bottom). 
Concretely, most of the FN samples correctly received a $\cflow_{hyp}\xrightarrow~0$.  
The plot shows, that $\cflow_{hyp}$ introduces some errors especially for higher \cflow estimates ($\cflow > \xi_1$), as there are some outliers around the bisecting line.
However, the corresponding bar plots depict that the errors are produced for a minority of samples. 
Thus, our evaluation indicates that the concept of $\mathfrak{B}_{hyp}$ generates accurate results for most FN as the results for $\cflow_{hyp}$ identified $\approx97\%$ of all FN that were also labeled as FN by means of $\cflow_{GT}$.

%
%

\section{Related Work}
\label{sec:relatedwork}

Although optical flow has been widely researched to enhance detection performance~\cite{Ramzan2016IntelligentPD,hu2017pushing,kang2017t}, the scope of our work is to exploit optical flow V\&V in the context of AD, \ie~we want to determine safety-critical faults in the DNN's output without performing particular architectural changes.

Two papers that focus on monitoring perception correctness for segmentation are~\cite{Varghese2020,zhang2022perceptual}.
The most similar approach to ours is proposed by Varghese \etal~\cite{Varghese2020} which presents a temporal consistency metric to measure the stability of consecutive semantic segmentation predictions utilizing optical flow.
Varghese~\etal~\cite{Varghese2020} argue that their approach may be used as a supplementary observer (online-monitor) to support safety requirements which also motivates our work.
Yet, the proposed metric weights all pixels equally whereas we focus on object detection, \ie~we are particularly interested in investigating the credibility of detections in safety-critical sequences.

Zhang~\etal\cite{zhang2022perceptual} propose a perceptual consistency measure between segmentation maps on consecutive image pairs to capture the temporal consistency of a video segmentation. 
Unlike optical flow, the perceptual consistency measure does not seek for exact pixel correspondences across two images but finds pairs of maximally correlated pixels to mitigate the impact of occlusion.
Our approach, however, utilizes optical flow to measure motion-based inconsistencies within bounding boxes.
This enables us to identify generally difficult cases for object detection like occlusion.

\section{Conclusion}
\label{sec:conclusions}
Safety-agnostic evaluation is a great safety concern for AD perception.
To consider safety in the evaluation of a pedestrian detector, this work leverages additional information extracted from sequences of images to inject domain-knowledge.
Particularly, it leverages temporal consistency information from optical flow to estimate the credibility of detections in a time-coherent pedestrian track.
Therefore, we introduce a novel metric \cflow to identify false negative predictions of a pedestrian detector.
To demonstrate the validity of \cflow, \textit{(i)} we perform controlled experiments with ground truth bounding boxes on a large AD dataset to show that \cflow achieves high accuracy with a low number of false alarms.
Furthermore, \textit{(ii)} we perform the computation of \cflow without ground truth based on the concept of hypothesized bounding boxes and demonstrate that even in the unsupervised case \cflow provides valid and very promising results.
This qualifies \cflow for perspective runtime applications such as an observer for safety-critical misdetections or active learning.

As future work, we see that the detection of non-credible samples can be further optimized, \eg~using optical flow vectors to refine the estimated hypothesized bounding boxes~\cite{true2021}.

%

%
%
\bibliographystyle{splncs04}
\bibliography{Literature}

\begin{thebibliography}{10}
\providecommand{\url}[1]{\texttt{#1}}
\providecommand{\urlprefix}{URL }
\providecommand{\doi}[1]{https://doi.org/#1}

\bibitem{Abrecht2023DeepLS}
Abrecht, S., Hirsch, A., Raafatnia, S., Woehrle, M.: Deep learning safety
  concerns in automated driving perception. ArXiv  \textbf{abs/2309.03774}
  (2023), \url{https://api.semanticscholar.org/CorpusID:261582607}

\bibitem{NTSB}
Board, N.T.S.: Collision between vehicle controlled by developmental automated
  driving system and pedestrian.
  \url{https://www.ntsb.gov/investigations/Pages/HWY18MH010.aspx} (18032018),
  accessed: 17.11.20232

\bibitem{nuscenes2019}
Caesar, H., Bankiti, V., Lang, A.H., Vora, S., Liong, V.E., Xu, Q., Krishnan,
  A., Pan, Y., Baldan, G., Beijbom, O.: {nuScenes: A} multimodal dataset for
  autonomous driving. CoRR  \textbf{abs/1903.11027} (2019),
  \url{http://arxiv.org/abs/1903.11027}

\bibitem{Ceccarelli2022EvaluatingO}
Ceccarelli, A., Montecchi, L.: Evaluating object (mis)detection from a safety
  and reliability perspective: Discussion and measures. IEEE Access
  \textbf{11},  44952--44963 (2022),
  \url{https://api.semanticscholar.org/CorpusID:252762677}

\bibitem{Argoverse}
Chang, M., Lambert, J., Sangkloy, P., Singh, J., Bak, S., Hartnett, A., Wang,
  D., Carr, P., Lucey, S., Ramanan, D., Hays, J.: Argoverse: {3D} tracking and
  forecasting with rich maps. CoRR  \textbf{abs/1911.02620} (2019),
  \url{http://arxiv.org/abs/1911.02620}

\bibitem{geissler}
Geissler, F., Qutub, S., Paulitsch, M., Pattabiraman, K.: A low-cost strategic
  monitoring approach for scalable and interpretable error detection in deep
  neural networks. In: Computer Safety, Reliability, and Security: 42nd
  International Conference, SAFECOMP 2023, Toulouse, France, September 20--22,
  2023, Proceedings. pp. 75--88. Springer-Verlag, Berlin, Heidelberg (2023).
  \doi{10.1007/978-3-031-40923-3_7},
  \url{https://doi.org/10.1007/978-3-031-40923-3_7}

\bibitem{hawkinsrichard}
Hawkins, R., Conmy, P.R.: Identifying run-time monitoring requirements for
  autonomous systems through the analysis of safety arguments. In:
  International Conference on Computer Safety, Reliability, and Security
  (2023), \url{https://api.semanticscholar.org/CorpusID:261894104}

\bibitem{hu2017pushing}
Hu, Q., Wang, P., Shen, C., van~den Hengel, A., Porikli, F.: Pushing the limits
  of deep cnns for pedestrian detection. IEEE Transactions on Circuits and
  Systems for Video Technology  \textbf{28}(6),  1358--1368 (2017)

\bibitem{kang2017t}
Kang, K., Li, H., Yan, J., Zeng, X., Yang, B., Xiao, T., Zhang, C., Wang, Z.,
  Wang, R., Wang, X., et~al.: T-cnn: Tubelets with convolutional neural
  networks for object detection from videos. IEEE Transactions on Circuits and
  Systems for Video Technology  \textbf{28}(10),  2896--2907 (2017)

\bibitem{argoverseHD}
Li, M., Wang, Y.X., Ramanan, D.: Towards streaming perception. In: Vedaldi, A.,
  Bischof, H., Brox, T., Frahm, J.M. (eds.) Computer Vision -- ECCV 2020. pp.
  473--488. Springer International Publishing, Cham (2020)

\bibitem{retina}
Lin, T., Goyal, P., Girshick, R.B., He, K., Doll{\'{a}}r, P.: Focal loss for
  dense object detection. In: {IEEE} International Conference on Computer
  Vision, {ICCV} 2017, Venice, Italy, October 22-29, 2017. pp. 2999--3007.
  {IEEE} Computer Society (2017). \doi{10.1109/ICCV.2017.324},
  \url{https://doi.org/10.1109/ICCV.2017.324}

\bibitem{LGH+21}
Lyssenko, M., Gladisch, C., Heinzemann, C., Woehrle, M., Triebel, R.: From
  evaluation to verification: Towards task-oriented relevance metrics for
  pedestrian detection in safety-critical domains. In: 2021 IEEE/CVF Conference
  on Computer Vision and Pattern Recognition Workshops (CVPRW). pp. 38--45
  (2021). \doi{10.1109/CVPRW53098.2021.00013}

\bibitem{LGH+22}
Lyssenko, M., Gladisch, C., Heinzemann, C., Woehrle, M., Triebel, R.: Towards
  safety-aware pedestrian detection in autonomous systems. In: International
  Conference on Intelligent Robots and Systems (IROS 2022). pp. 293--300.
  {IEEE} (2022). \doi{10.1109/IROS47612.2022.9981309}

\bibitem{Nishimura2021SDOFTrackerFA}
Nishimura, H., Komorita, S., Kawanishi, Y., Murase, H.: {SDOF-Tracker:} fast
  and accurate multiple human tracking by skipped-detection and optical-flow.
  CoRR  \textbf{abs/2106.14259} (2021), \url{https://arxiv.org/abs/2106.14259}

\bibitem{Ramzan2016IntelligentPD}
Ramzan, H., Fatima, B., Shahid, A., Ziauddin, S., Ali, A.: Intelligent
  pedestrian detection using optical flow and hog. International Journal of
  Advanced Computer Science and Applications  \textbf{7} (01 2016).
  \doi{10.14569/IJACSA.2016.070955}

\bibitem{siefke}
Siefke, L., Sommer, V., Baylan, M.C., Grunske, L.: Probabilistic spatial
  relations for monitoring behavior of road users. In: Computer Safety,
  Reliability, and Security: 42nd International Conference, SAFECOMP 2023,
  Toulouse, France, September 20--22, 2023, Proceedings. pp. 151--164.
  Springer-Verlag, Berlin, Heidelberg (2023).
  \doi{10.1007/978-3-031-40923-3_12},
  \url{https://doi.org/10.1007/978-3-031-40923-3_12}

\bibitem{Teed2020RAFTRA}
Teed, Z., Deng, J.: {RAFT:} recurrent all-pairs field transforms for optical
  flow. CoRR  \textbf{abs/2003.12039} (2020),
  \url{https://arxiv.org/abs/2003.12039}

\bibitem{true2021}
True, J., Khan, N.: Motion vector extrapolation for video object detection.
  CoRR  \textbf{abs/2104.08918} (2021), \url{https://arxiv.org/abs/2104.08918}

\bibitem{Varghese2020}
Varghese, S., Bayzidi, Y., Bär, A., Kapoor, N., Lahiri, S., Schneider, J.D.,
  Schmidt, N., Schlicht, P., Hüger, F., Fingscheidt, T.: Unsupervised temporal
  consistency metric for video segmentation in highly-automated driving. In:
  2020 IEEE/CVF Conference on Computer Vision and Pattern Recognition Workshops
  (CVPRW). pp. 1369--1378 (2020). \doi{10.1109/CVPRW50498.2020.00176}

\bibitem{WolfDE21}
Wolf, M., Douat, L.R., Erz, M.: Safety-aware metric for people detection. In:
  24th {IEEE} International Intelligent Transportation Systems Conference,
  {ITSC} 2021, Indianapolis, IN, USA, September 19-22, 2021. pp. 2759--2765.
  {IEEE} (2021). \doi{10.1109/ITSC48978.2021.9564734},
  \url{https://doi.org/10.1109/ITSC48978.2021.9564734}

\bibitem{zhang2022perceptual}
Zhang, Y., Borse, S., Cai, H., Wang, Y., Bi, N., Jiang, X., Porikli, F.:
  Perceptual consistency in video segmentation. In: {IEEE/CVF} Winter
  Conference on Applications of Computer Vision, {WACV} 2022, Waikoloa, HI,
  USA, January 3-8, 2022. pp. 2623--2632. {IEEE} (2022).
  \doi{10.1109/WACV51458.2022.00268},
  \url{https://doi.org/10.1109/WACV51458.2022.00268}

\end{thebibliography}

\end{document}